%% file: main.tex
\PassOptionsToPackage{dvipsnames,table}{xcolor}
\documentclass[sigconf]{acmart}
\usepackage{algorithm}
\usepackage{algorithmic}

\usepackage[ruled,vlined,algo2e]{algorithm2e}
\usepackage{setspace}
\usepackage{multirow}
\usepackage{cleveref}
\usepackage{enumitem}
\usepackage{threeparttable}
\usepackage{graphicx}
\usepackage{caption}
\usepackage{subcaption}
\usepackage{adjustbox} 
\usepackage{makecell}
\usepackage{booktabs}
\usepackage{fontawesome}
\usepackage{amsmath}
\usepackage{microtype}
\usepackage{amsfonts}
\usepackage{MnSymbol}
\usepackage{bbding} 

\AtBeginDocument{%
  }

\setcopyright{acmlicensed}
\copyrightyear{2018}
\acmYear{2018}
\acmDOI{XXXXXXX.XXXXXXX}
\acmConference[Conference acronym 'XX]{Make sure to enter the correct
  conference title from your rights confirmation email}{June 03--05,
  2018}{Woodstock, NY}
\copyrightyear{2025}
\acmYear{2025}
\setcopyright{acmlicensed}
\acmConference[CIKM '25] {Proceedings of the 34th ACM International Conference on Information and Knowledge Management}{ November 10--14, 2025}{Seoul, Republic of Korea.}
\acmBooktitle{Proceedings of the 34th ACM International Conference on Information and Knowledge Management (CIKM '25), November 10--14, 2025, Seoul, Republic of Korea}
\acmISBN{979-8-4007-2040-6/2025/11}
\acmDOI{10.1145/3746252.3761580}

\newcommand{\modelname}{\textbf{MAL} } 
\newcommand{\acronym}[1]{\underline{\textbf{#1}}}

\newcommand{\up}[1]{\textcolor{OliveGreen}{\footnotesize \ $\uparrow${#1\%}}}

\newcommand{\basex}[1]{\textcolor{gray!50}{\footnotesize \ $\uparrow${#1\%}}}
\newcommand{\basexx}[1]{\textcolor{gray!85}{\footnotesize \ $\uparrow${#1\%}}}

\makeatletter

\newcommand{\Rmnum}[1]{\expandafter\@slowromancap\romannumeral #1@}
\makeatother

\begin{document}

\title{See Beyond a Single View: Multi-Attribution Learning Leads to Better Conversion Rate Prediction} 



\author{Sishuo Chen}
\authornote{Equal contribution.} 
\email{chensishuo.css@alibaba-inc.com}
\author{Zhangming Chan} \authornotemark[1]
\email{zhangming.czm@alibaba-inc.com}
\author{Xiang-Rong Sheng}
\email{xiangrong.sxr@alibaba-inc.com}
\affiliation{
   \institution{Taobao \& Tmall Group of Alibaba}
   \city{Beijing}
   \country{China}
}

\author{Lei Zhang}
\email{zl165646@alibaba-inc.com}
\author{Sheng Chen}
\email{chensheng.cs@alibaba-inc.com}
\author{Chenghuan Hou}
\email{jinyao@alibaba-inc.com}
\affiliation{
   \institution{Taobao \& Tmall Group of Alibaba}
   \city{Beijing}
   \country{China}
}

\author{Han Zhu}
\email{zhuhan.zh@alibaba-inc.com}
\author{Jian Xu}
\email{xiyu.xj@alibaba-inc.com}
\author{Bo Zheng}\authornote{Corresponding author.} 
\email{bozheng@alibaba-inc.com}
\affiliation{
   \institution{Taobao \& Tmall Group of Alibaba}
   \city{Beijing}
   \country{China}
}



\renewcommand{\shortauthors}{Chen and Chan et al.}

\begin{abstract}

Conversion rate (CVR) prediction is a core component of online advertising systems, where the \textit{attribution mechanisms}---rules for allocating conversion credit across user touchpoints---fundamentally determine label generation and model optimization.
While many industrial platforms support diverse attribution mechanisms (e.g., First-Click, Last-Click, Linear, and Data-Driven Multi-Touch Attribution), 
conventional approaches restrict model training to labels from a single production-critical attribution mechanism, discarding complementary signals in alternative attribution perspectives. 

To address this limitation, we propose a novel \acronym{M}ulti-\acronym{A}ttribution \acronym{L}earning (\textbf{MAL}) framework for CVR prediction that integrates signals from multiple attribution perspectives to better capture the underlying patterns driving user conversions. 
Specifically, \modelname is a joint learning framework consisting of two core components: the Attribution Knowledge Aggregator (\textbf{AKA}) and the Primary Target Predictor (\textbf{PTP}). AKA
is implemented as a multi-task learner that integrates knowledge extracted from diverse attribution labels. PTP, in contrast, focuses on the task of generating well-calibrated conversion probabilities that align with the system-optimized attribution metric (e.g., CVR under the Last-Click attribution), ensuring direct compatibility with industrial deployment requirements.
Additionally, we propose \text{CAT}, a novel training strategy that leverages the Cartesian product of all attribution label combinations to generate enriched supervision signals. This design substantially enhances the performance of the attribution knowledge aggregator. 
Empirical evaluations demonstrate the superiority of \modelname over single-attribution learning baselines, achieving +0.51\% GAUC improvement on offline metrics. Online experiments demonstrate that \modelname achieved a +2.6\% increase in ROI (Return on Investment).

\end{abstract}


\maketitle

\input{chapters/intro}

\input{chapters/related_work}

\input{chapters/method}

\input{chapters/experiments}

\input{chapters/conclusion}

\section*{GenAI Usage Disclosure}
No generative AI software was used in the writing of this paper.

\bibliographystyle{ACM-Reference-Format}
\bibliography{acmart}

\end{document}

%% file: chapters/intro.tex
\section{Introduction}

\begin{figure}[!t]
\centering 
\includegraphics[width=\columnwidth]{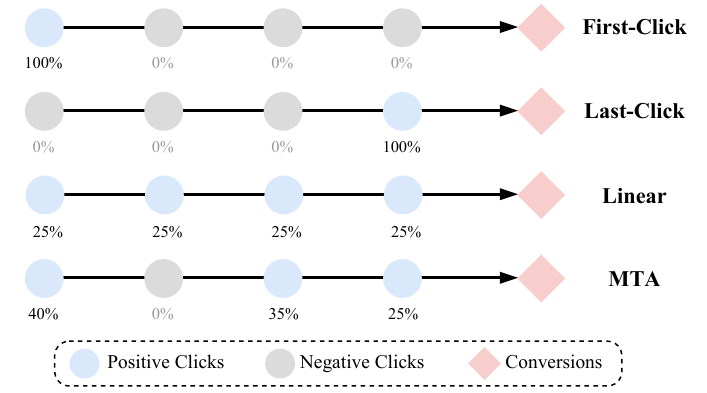}
\caption{Illustration of different attribution mechanisms.} 
\label{fig:attr_view} 
\end{figure}

In online advertising, advertisers compete for traffic resources through auction mechanisms to maximize conversion outcomes (e.g., purchases)~\cite{evans2009online,goldfarb2011online}. To establish causal relationships between advertising interactions and user conversions, industrial systems rely on \textit{attribution mechanisms}---formalized rules that quantify the contribution of each touchpoint (typically operationalized as ad clicks) to the eventual conversion event. These mechanisms systematically allocate credit across historical advertising interactions, enabling performance evaluation and campaign optimization. 
Several predominant attribution mechanisms are used in industrial practice, including: 
\begin{itemize}[leftmargin=*]
    \item \textbf{Last-Click Attribution}: Concentrates 100\% attribution weight on the final touchpoint preceding conversion.
    \item \textbf{First-Click Attribution}: Attributes full credit exclusively to the initial user interaction.
    \item \textbf{Linear Attribution}: Distributes weights uniformly across all touchpoints. 
    \item \textbf{Data-Driven Multi-Touch Attribution (MTA)}: Learns weight allocation through causal inference models~\cite{zhou2019deep,yao2022causalmta,kumar2020camta,bencina2025lidda}.
\end{itemize} 
Figure~\ref{fig:attr_view} illustrates the characteristics of these mechanisms. 

Although many online advertising platforms offer diverse attribution mechanisms for advertisers to evaluate campaign effectiveness, most systems optimize exclusively for a single attribution metric, such as the Last-Click metric, due to operational simplicity. Specifically, conversion labels generated by this system-optimized attribution mechanism are used to train conversion rate (CVR) prediction models. These predicted values then guide automated bidding~\cite{yang2019bid,wang2017display,cai2017real,he2021unified,jin2018real,mou2022sustainable} and traffic allocation~\cite{zhu2017optimized,borgs2007dynamics,liu2021neural,fan2025two,wang2022designing,zhu2024contextual,chen2022hierarchically}. 

This study systematically examines how different attribution mechanisms impact the performance of CVR prediction models. Our investigation addresses a critical gap in the existing literature: while most prior works~\cite{chapelle2014modeling,lu2017practical,esmm,defer,defuse,zhuang2025practice,chan2023capturing,chan2020selection} confine model training to labels
from a single system-optimized attribution mechanism, 
the potential of integrating multiple attribution perspectives remains underexplored. 
This oversight carries significant practical implications. When models are trained under this single-attribution paradigm, they fail to leverage signals from alternative attribution perspectives that could improve the modeling of user intent. For instance, under last-click attribution, all nonterminal touchpoints, even those that may have played an influential role in early stages of the journey, are assigned the same negative label regardless of their actual impact on conversion. As a result, such approaches disregard potentially valuable behavioral patterns that reflect meaningful user interactions. 


To address this limitation, we propose a novel \acronym{M}ulti-\acronym{A}ttribution \acronym{L}earning (\textbf{MAL}) framework that integrates multi-attribution signals to enhance prediction accuracy for the target attribution mechanism. The theoretical basis of multi-attribution learning lies in the complementary insights offered by different attribution mechanisms, which together enable a more comprehensive understanding of user conversion behavior.  Specifically, we formalize CVR prediction under the system-optimized attribution mechanism as the primary optimization objective, while constructing a joint learning architecture that incorporates conversion signals from multiple attribution mechanisms---including first-click, last-click, linear, and MTA. The proposed framework aim to leverages the diverse attribution signals to refine the primary target estimation while ensures deployment compatibility with industrial systems.

Concretely, our approach follows a joint learning manner consisting of two core components: (1) \textbf{Attribution Knowledge Aggregator (AKA)}, implemented as a multi-task learner that extracts knowledge from diverse attribution labels; and (2) \textbf{Primary Target Predictor (PTP)}, which leverage the extracted knowledge to enhance the main conversion prediction task. Specifically, AKA jointly predicts CVR under multiple attribution mechanisms by employing dedicated prediction towers for each attribution mechanism. From the outputs of these towers, we extract high-level conversion knowledge representations, which capture the pattern of user conversion behavior. PTP then fuses these representations with the original features to achieve improved CVR prediction performance under the system-optimized attribution mechanism (primary target). 

To further enhance knowledge transfer, we introduce CAT, which is short for Cartesian-based Auxiliary Training, a novel auxiliary task training strategy based on combinations of cross-attribution labels. It generates multi-dimensional supervision signals by computing the Cartesian product of attribution labels across multiple mechanisms. This approach goes beyond traditional single-mechanism formulations by producing fine-grained supervisory signals that capture more nuanced aspects of user conversion behavior.

We summarize our main contributions as follows:
\begin{itemize}[leftmargin=*]
    \item  We propose a novel multi-attribution learning framework that integrates multiple attribution signals to enhance performance under the system-optimized attribution mechanism. This work pioneers the industrial deployment of multi-attribution learning in CVR modeling, addressing critical limitations in existing literature that focus exclusively on single-attribution learning.
    \item We propose an implementation of the proposed MAL framework, which incorporates the AKA and PTP architectures and the Cartesian product of attribution labels across different mechanisms. These components collectively enable more comprehensive modeling of user conversion intentions. 
    \item We conduct comprehensive experiments on the display advertising system in Alibaba, demonstrating statistically significant improvements: +2.6\% in ROI during online A/B tests compared to the production baseline. We believe that the insights we have gathered will serve as valuable resources for industrial practitioners seeking to leverage multi-attribution learning to enhance the CVR prediction.
\end{itemize}

%% file: chapters/related_work.tex
\section{Related Work}

\subsection{Conversion Rate Prediction} \label{subsec:cvr}

Conversion rate (CVR) prediction~\cite{chapelle2014modeling,lu2017practical} plays an important role in online advertising.
Researchers have made extensive efforts to improve CVR prediction models from various perspectives, such as delay feedback modeling~\cite{yang2021capturing,defer,defuse}, model architecture design~\cite{zhuang2025practice}, and joint learning with other tasks~\cite{esmm,wang2022escm2,zhao2023ecad}.
Although the attribution mechanism decides the learning targets of CVR prediction models, existing studies~\cite{chapelle2014modeling,lu2017practical,esmm,defer,defuse,zhuang2025practice,yang2021capturing,wang2022escm2,zhao2023ecad,chan2023capturing} are conducted under a single attribution mechanism, ignoring valuable auxiliary signals underlying labels generated from alternative attribution perspectives.
To our knowledge, our work takes the first step to explore the \textit{multi-attribution learning} framework and mine learning signals from labels derived by diverse attribution mechanisms.~\looseness=-1

\subsection{Attribution Mechanisms} \label{subsec:mta} 

Attribution modeling serves as an indispensable module in online advertising for systematically evaluating the influence of every user touchpoint across conversion pathways~\cite{gaur2020attribution}.
Existing attribution mechanisms can be categorized into two primary groups: \textbf{(1) rule-based attribution} following heuristic rules for credit assignment, such as last-click, first-click, time-decay, and linear attribution~\cite{berman2018beyond};
\textbf{(2) data-driven multi-touch attribution} mechanisms based on attribution models~\cite{shao2011data} relying on causal inference techniques, such as DeepMTA~\cite{zhou2019deep}, CAMTA~\cite{kumar2020camta}, CausalMTA~\cite{yao2022causalmta}, and LiDDA~\cite{bencina2025lidda}. Our study opens up a new avenue for enhancing CVR prediction with labels generated under different attribution mechanisms. 

\subsection{Multi-Task Learning for Recommendation}
\label{subsec:mtl}

The multi-task learning (MTL) paradigm~\cite{zhang2021survey} aims to build models capable of predicting multiple targets simultaneously in recommendation systems~\cite{caruana1997multitask,sheng2021one} and other applications~\cite{chapelle2014modeling,misra2016cross,fan2017hd}.
A wide array of MTL studies for recommendation focus on general architecture design, so that parameter learning for different tasks could benefit from each other, such as MMoE~\cite{mmoe}, PLE~\cite{PLE}, AdaTT~\cite{li2023adatt}, and HoME~\cite{wang2024home}.
Another line of works focuses on learning paradigm design for related recommendation tasks, such as ESMM~\cite{esmm} and ESCM-2~\cite{wang2022escm2} for joint learning of CTR estimation and CVR prediction and HM$^3$~\cite{wen2021hierarchically} for CVR prediction and micro and behavior modeling.
The multi-attribution learning for CVR prediction could also be regarded as an MTL problem, for which we will investigate suitable model structure and learning targets in this research.


%% file: chapters/method.tex
\section{Methods}

\subsection{Preliminaries}
\label{sec:prelimiary}

\begin{figure}[t]
\centering 
\includegraphics[width=\columnwidth]{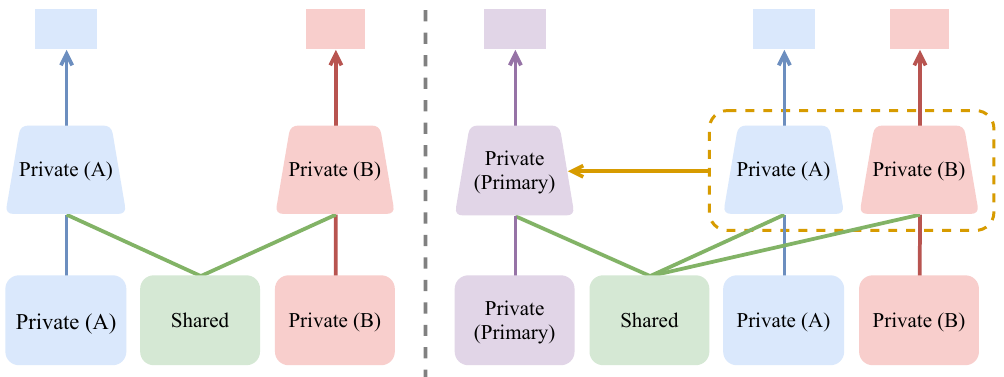}
\caption{Mechanism comparison between traditional multi-task learning approaches (left) and our MAL framework (right). The primary target is selected from options A or B based on business requirements in the MAL framework.} 
\label{fig:model_compare} 
\end{figure}

Although industrial online advertising systems often provide advertisers with performance reports under multiple attribution methods, they predominantly optimize conversion outcomes using a single production-critical mechanism due to operational simplicity. The labels generated from this mechanism (e.g., Last-Click or MTA) are referred to as the \textbf{primary target} in this work. More specifically, advertising systems utilize these primary target labels to train CVR prediction models, and the predicted conversion probabilities subsequently inform automated bidding and traffic allocation. In this work, we maintain this setup to ensure the practical applicability of model optimization in production systems, while our primary focus remains on improving the prediction performance of the CVR model with respect to the primary target.

\begin{figure*}[t]
\centering 
\includegraphics[width=0.8\textwidth]{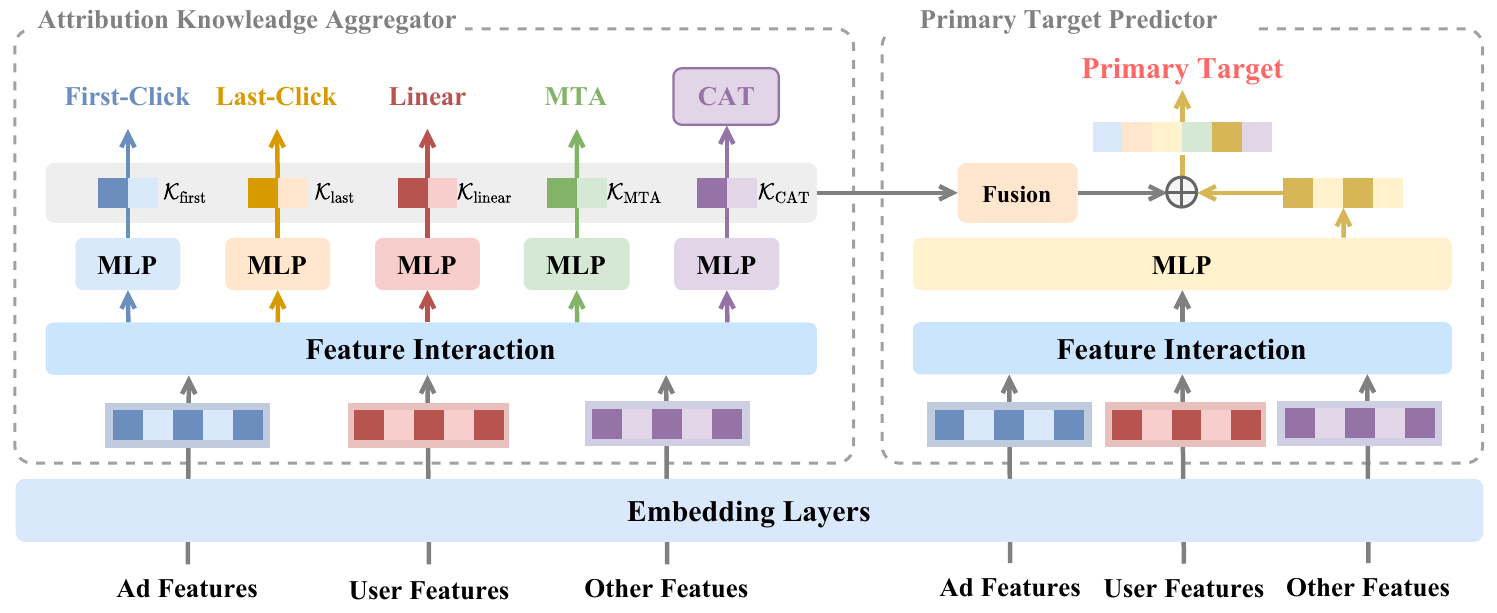}
\caption{The overall framework of \modelname.} 
\label{fig:overall} 
\end{figure*}

Besides the primary target, our approach differs from existing CVR prediction methods in that it leverages conversion signals from multiple attribution perspectives. Concretely, our Multi-Attribution Learning (\text{MAL}) framework incorporates these multi-attribution signals during training and enhances the prediction performance on the primary target through a joint learning paradigm. 

Formally, we frame CVR prediction with labels from diverse attribution mechanisms as a weighted binary classification problem with the following sample definition: 
\begin{itemize}[leftmargin=*] 
    \item \textbf{Positive Samples.} Samples with non-zero attribution weights constitute positive samples, with attribution weights serving as sample importance weights.
    \item \textbf{Negative Samples.} Samples receiving zero attribution weights are treated as negative samples. 
\end{itemize} 
In this study, we utilize four attribution mechanisms: last-click, first-click, linear, and multi-touch attribution (MTA). Importantly, the attribution methods employed in \modelname are not limited to these four - the framework supports seamless integration of additional attribution mechanisms without architectural constraints.

\input{tables/pos_ratio}

Notably, as the Linear and MTA attribution methods distribute conversions to multiple clicks, they naturally produce more positive samples under the aforementioned weighted binary classification setup. Table~\ref{tab:pos_ratio} presents the proportions of positive samples under these two attribution methods collected from the display advertising platform of Taobao, with Last-Click serving as the baseline. 
The Linear and MTA methods yield 42\% and 15\% more positive samples, respectively, than the last-click method. This comparison reveals the potential information gains inherent in diverse attribution strategies. Motivated by this insight, we propose a novel model architecture and an effective formulation of auxiliary learning objectives, as detailed below.

\subsection{Overall Workflow of \modelname}
The goal of MAL is to systematically integrate labels from multiple attribution mechanisms to enhance prediction performance for the primary target. To achieve this goal, we structure MAL with two core components: the Attribution Knowledge Aggregator (\textbf{AKA}) and the Primary Target Predictor (\textbf{PTP}), as illustrated in Figure~\ref{fig:overall}. 

In the first component, AKA serves as a multi-task learner. Specifically, it is trained on all attribution labels through a joint multi-task learning paradigm, enabling it to capture a more comprehensive understanding of user conversion patterns. Moreover, we propose Cartesian-based Auxiliary Training (\textbf{CAT}), a novel training strategy that leverages the Cartesian product of all attribution label combinations to generate enriched supervision signals. This design significantly enhances the performance of AKA (details in \S~\ref{subsec:aka}). 

The second component, PTP, functions similarly to conventional CVR prediction models and is specifically designed for predicting the primary target. Notably, PTP incorporates the knowledge learned by AKA to further improve its performance on the primary targets (details in \S~\ref{subsec:ptp}). Overall, AKA and PTP are implemented under a shared-bottom paradigm, which helps reduce deployment costs while maintaining modeling effectiveness. Moreover, Figure~\ref{fig:model_compare} highlights the key distinctions between traditional multi-task learning approaches and the proposed MAL framework. 

\subsection{Component \Rmnum{1}: Attribution Knowledge Aggregator} 
\label{subsec:aka} 

\subsubsection{Architecture} 
As illustrated in Figure~\ref{fig:overall}, the Attribution Knowledge Aggregator (AKA) is a multi-task learning module constructed upon a standard CVR prediction architecture. Specifically, AKA processes three types of input features: ad features, user features, and contextual features. First, an embedding layer maps these discrete/categorical features into a dense, low-dimensional embedding space. Subsequently, feature interaction modules---such as attention-based networks---are employed to capture cross-feature interactions between ad-user pairs. The output representations from these modules are aggregated into a shared representation vector $v$, which serves as the foundation for the multi-task prediction. For each attribution label, we design a dedicated prediction tower (i.e., fully-connected layers) that takes this shared representation $v$ as input and independently predicts the conversion probability under the corresponding attribution labels. 

\subsubsection{Multi-Task Optimization} 
\label{sec:mto}

AKA is a multi-task learner that simultaneously predicts CVR under diverse attribution labels.
As described earlier, AKA adopts a shared-bottom architecture, where multiple labels jointly supervise the model to capture complementary signals from different attribution perspectives. Note that although we employ only the standard shared-bottom structure, other multi-task learning methods are also compatible with AKA in principle. 
The training data $\mathcal{D}$ comprises user-ad interactions ($u$, $i$), each associated with four conversion labels generated under one of four attribution rules: $l_\text{last}$ from Last-Click, $l_\text{first}$ from First-Click, $l_\text{linear}$ from Linear, and $l_\text{mta}$ from MTA. Notably, all labels $l$ take a value in the range [0,1], reflecting the partial credit assigned to the corresponding click under each attribution mechanism. 

For each interaction ($u$, $i$), AKA is trained to predict the likelihood of conversion under a specific attribution method. This is formulated as a weighted binary classification problem, where the model minimizes the weighted cross-entropy loss between its predicted probability and the binary label $l$ and sample weight $w$. Formally, for a given attribution mechanism $a$, the loss is defined as:
\begin{equation}
\begin{aligned}
\label{eq:aka-cr} 
\mathcal{L}_{a} = &\sum_{\mathcal{D}} w \cdot \left[ (-l \cdot \log \sigma(\hat{y}) - (1-l) \cdot \log (1-\sigma(\hat{y})) \right], 
\end{aligned} 
\end{equation} 
where $\hat{y}$ denotes the output logit, which is transformed into a conversion probability through the sigmoid function $\sigma(\cdot)$. The prediction objective under each attribution method follows the same formulation, differing only in the corresponding supervision signal. 

\subsubsection{Conversion Knowledge Representation} 
\label{sec:ckr}
We extract the vectors from the penultimate layer outputs of the prediction towers, deriving four distinct \textit{conversion knowledge vectors}: $\mathcal{K}^{\text{first}}$, $\mathcal{K}^{\text{last}}$, $\mathcal{K}^{\text{linear}}$, and $\mathcal{K}^{\text{mta}}$. These vectors correspond to specific attribution mechanisms (first-click, last-click, linear, and MTA) and are derived from their respective prediction towers. 
Finally, the four conversion knowledge vectors are concatenated into a unified knowledge embedding $\mathcal{K}$, serving as the final conversion knowledge input for predicting the primary target.

\subsection{Component \Rmnum{2}: Primary Target Predictor} 
\label{subsec:ptp} 
Building upon the rich conversion knowledge representation $\mathcal{K}$, we construct the Primary Target Predictor (PTP), which serves as the main CVR prediction module for the primary target. PTP retains the same architecture as our production CVR prediction model to ensure compatibility with existing infrastructure and deployment pipelines. It also adopts a shared-bottom design in conjunction with AKA, facilitating efficient deployment into industrial systems. The key distinction between PTP and the production model lies in the incorporation of external conversion knowledge: while the production model operates solely on raw input features, PTP further utilizes the knowledge vector $\mathcal{K}$ extracted from AKA, enabling more informed predictions for the primary target. 

Concretely, PTP first employs an embedding layer, feature interaction modules, and a shallow MLP to map raw inputs into an intermediate vector $v_{p}\in\mathbb{R}^d$.
Motivated by the design principles of KEEP~\cite{zhang2022keep}, we propose a compact knowledge fusion mechanism within PTP to integrate the knowledge vector $\mathcal{K}$ with  $v_{p}$.
Specifically, we project $\mathcal{K}$ into the semantic space of $v_{p}$ via an MLP:
\begin{equation}
    v_{a} = \text{MLP}(\mathcal{K}),
\end{equation}
where $v_{a}$ denotes the aligned knowledge vector. Subsequently,  $v_{a}$ and $v_{p}$ are fused through element-wise addition, yielding an enriched representation $v_{\text{fusion}}=v_p+v_a$. 

This fused representation serves as input to a binary classification head for predicting the primary target label. Importantly, PTP is trained using its cross-entropy objective as Eq.~\eqref{eq:aka-cr}, ensuring that the injected conversion knowledge enhances---not distorts---the model’s alignment with the primary target. This end-to-end training strategy allows PTP to selectively leverage multi-attribution signals while maintaining precision for the main target.

\input{tables/targets_combo_effect}

\subsection{Cartesian-based Auxiliary Training} 
\label{subsec:target} 
To further enrich the supervision signal in MAL and capture more comprehensive behavioral patterns, we propose \textbf{CAT} (Cartesian-style Auxiliary Target), which integrates higher-order interactions among attribution signals in AKA.

Before delving into methodological details, we conduct exploratory experiments under various attribution label combinations using the proposed MAL framework. Table~\ref{tab:targets_effects} reports the performance improvements on the primary target (Last-Click).
The results demonstrate that incorporating all available conversion labels yields the largest performance gains over AKA trained with a single learning objective.  This observation supports our hypothesis that \textbf{a more diverse set of conversion labels encourages AKA to capture more comprehensive behavioral patterns, ultimately leading to greater improvements on the primary target.}

\begin{figure}[t]
\centering 
\includegraphics[width=0.75 \columnwidth]{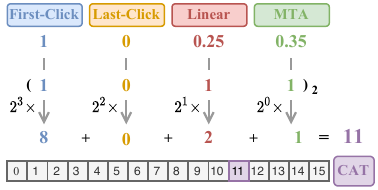}
\caption{A running example of the CAT label generation.} 
\label{fig:cat_label} 
\end{figure}

Inspired by this observation, we propose CAT, a novel training strategy designed to further enrich the supervision signals derived from multiple attribution mechanisms.  Specifically, CAT constructs a multi-class classification task based on the Cartesian product of N binary conversion targets generated under different attribution methods.  This leads to a composite label space containing $2^N$  combinations, where each class represents a distinct pattern of conversion outcomes across all attribution views.  Algorithm~\ref{alg:cat} outlines the process for generating CAT labels, and Figure \ref{fig:cat_label} provides an example.  In our industrial setting, where four attribution mechanisms are considered, CAT becomes a 16-class classification problem.

From a modeling perspective, within the MAL framework described in \S~\ref{sec:mto}, we introduce an auxiliary prediction tower to fit the CAT objective, as illustrated in the left half of Figure~\ref{fig:overall}. The conversion knowledge vector $\mathcal{K}_\text{CAT}$, constructed following the same formulation as described in \S~\ref{sec:ckr}, is then extracted from this tower and incorporated into the overall conversion knowledge representation $\mathcal{K}$ learned by AKA. This integration further enhances the model's ability to capture complex user behavior patterns from diverse attribution perspectives.

\begin{algorithm}
\caption{Compute the \textbf{CAT} Label}\label{alg:cat}
\begin{algorithmic}[1]
\REQUIRE Array $\mathbf{A}$ of length $N$ with elements in \{0, 1\}, namely the labels under $N$ types of attribution mechanisms
\ENSURE The CAT label $\mathbf{O}$  $\in [0, 2^N - 1]$

\STATE Initialize $\mathbf{O} \gets 0$
\FOR{$i=0$ to $N-1$}
\STATE $\mathbf{O} \gets \mathbf{O} + \mathbf{A}[i] \times 2^{i}$
\ENDFOR

\end{algorithmic}
\end{algorithm}


%% file: tables/pos_ratio.tex
\begin{table}[t] 
\centering
\caption{CVR positive sample percentages under different attribution mechanisms, where the ratio under the last-click attribution mechanism is represented as $r$ for confidentiality.} 
\begin{tabular}{@{}l|ccc@{}}
\toprule
& \textbf{Last-Click} & \textbf{Linear} & \textbf{MTA}  \\
\midrule
Positive Ratio & $r$ & 1.42$r$ &  1.15$r$ \\ 
\bottomrule
\end{tabular} 
\label{tab:pos_ratio}
\end{table} 

%% file: tables/targets_combo_effect.tex
\begin{table}[t]
\centering
\caption{The relative increase in AUC and GAUC metrics when different attribution methods are leveraged.}
\resizebox{0.35\textwidth}{!}{ 
\begin{tabular}{@{}l|cc@{}}
\toprule
\textbf{Attribution Methods}  & $\mathbf{\Delta}$\textbf{GAUC} & $\mathbf{\Delta}$\textbf{AUC} \\ \midrule
First-Click                  &  +0.18\%  &  +0.07\%  \\
Linear                       &  +0.25\%  &  +0.09\%  \\
MTA                          &  +0.36\%  &  +0.09\%  \\
First-Click \& Linear \& MTA &  \textbf{+0.38\%}  &  \textbf{+0.14\%}  \\ 
\bottomrule
\end{tabular}}
\label{tab:targets_effects}
\end{table}

%% file: chapters/experiments.tex
\section{Experiments}
\subsection{Experimental Setup}

\subsubsection{Dataset}

Given the absence of publicly available CVR prediction datasets containing labels from multiple attribution mechanisms, we conduct experiments on real-world data collected from the display advertising system of Taobao, Alibaba.
The training dataset is derived from user click logs of two months, and the logs of the following day serve as the test set. 
In this dataset, labels under four kinds of attribution mechanism are available, namely last-click, first-click, linear, and MTA.

To validate the generality of the proposed MAL framework, we conduct two groups of experiments where the last-click label and the MTA label serve as the primary target, respectively.

\subsubsection{Compared Methods}
We compare \modelname against the production baseline  (\textbf{Base}) and multi-task baselines. 
The  \textbf{Base} model is our well-developed production model implementing single-attribution learning.  The model involves hybrid architectures combining the SimTier architecture for multimodal feature extraction~\cite{sheng2024enhancing}, attention modules for user behavior sequence feature extraction~\cite{din}, an MLP backbone for feature interaction, and a prediction head outputting the primary target. 
As for the multi-task baselines, we implement state-of-the-art multi-task learning approaches  where each auxiliary target corresponds to an independent prediction task:~\looseness=-1
\begin{itemize}[leftmargin=*]
    \item \textbf{Shared-Bottom~\cite{caruana1997multitask}.}  Shared-Bottom is a multi-task model that shares bottom layers. In our implementation, the embedding layer and the first three fully-connected layers are shared across tasks.~\looseness=-1
    \item \textbf{MMoE~\cite{mmoe}.} MMoE implicitly models task relationships in multi-task learning, especially when tasks have distinct label spaces. It employs shared experts and learns task-specific gating networks to dynamically combine expert outputs based on input features.
    \item \textbf{PLE~\cite{PLE}.} PLE extends MMoE by introducing a shared expert for global knowledge and a task-specific expert for each task, enhancing the fusion of general and specialized information. It further adopts a two-layer stacked structure to better learn both shared and task-specific representations.
    \item \textbf{HoME~\cite{wang2024home}.} HoME improves the MoE framework by introducing three structural enhancements: Expert normalization combined with Swish activation~\cite{ramachandran2017searching}, a hierarchical masking mechanism, and feature-gating and self-gating modules. We implement HoME under the same setup as PLE, ensuring fair comparison.
\end{itemize} 
All methods are trained with one epoch~\cite{ZhangSZJHDZ2022OneEpoch}.

\subsubsection{Evaluation Metrics} 
We report the widely-used \textbf{AUC} and \textbf{GAUC} (Group AUC) metrics~\cite{schutze2008introduction,din,ShengGCYHDJXZ2023JRC,BianWRPZXSZCMLX2022CAN}  of the primary task for assessing model performance. 
AUC is a widely used metric for measuring how well a model can distinguish between the distributions of positive and negative samples, making it particularly suitable for CVR prediction. In the practice, we incorporate sample weights into the computation of the AUC metric.
GAUC reflects the model's ability to rank conversion samples within the same user group and serves as the core evaluation criterion in our production system. It has also been empirically validated to exhibit stronger consistency with online performance. The definition of GAUC is given in Eq.~\eqref{eqn:gauc}:
\begin{equation}
\small 
\label{eqn:gauc} 
\begin{aligned} 
    \textrm{GAUC} = \frac{\sum_{u=1}^U \# \textrm{click}(u) \times \textrm{AUC}_u}{\sum_{u=1}^U \# \textrm{click}(u)},
\end{aligned}
\end{equation}
in which $U$ represents the number of users, $\#\textrm{click}(u)$ denotes the number of clicks for the $u$-th user, and $\textrm{AUC}_u$ is the AUC computed using the samples from the $u$-th user. 

\subsection{Experimental Results}

\subsubsection{Overall Performance} 

We display the ranking performance of \modelname and baselines in Table~\ref{tab:main_results} and make two key observations.

\textbf{First, \modelname outperforms the production baseline (Base) by significant margins, demonstrating the advantage of multi-attribution learning}. 
Specifically, when the last-click labels serve as the primary target, \modelname surpasses Base by 0.51\% in GAUC and 0.14\% in AUC; when the MTA labels are the primary targets, \modelname exceeds Base by 0.75\% in GAUC and 0.21\% in AUC. 
It is noteworthy that improvements of 0.5\% in GAUC in the offline evaluation are significant enough to bring a significant increase in online revenue for our advertising platform.

\textbf{Second, \modelname beats competing multi-task learning methods, showing its efficiency in leveraging diverse attribution labels.} 
Concretely, \modelname exceeds the previous state-of-the-art (SOTA) model by 0.20\% and 0.15\% in GAUC  when the last-click labels and the MTA labels are primary targets, respectively. 
Regarding AUC, \modelname surpasses the previous SOTA model by 0.05\% and 0.06\% in the two groups of experiments, respectively.
The outstanding performance of \modelname proves that vanilla adoption of MTL methods is suboptimal for multi-attribution learning, and substantial improvements can be achieved by careful design of model architecture and optimization strategies, like our AKA module and CAT strategy. 

\input{tables/main_results}

\subsubsection{Ablation Studies} 

To verify the contribution of the two key designs in \textbf{MAL}, namely the AKA module and the CAT target, we conduct ablation experiments.
As shown in Table~\ref{tab:label_ablation}, the AKA architecture alone brings a GAUC increase of 0.38\% and an AUC lift of 0.14\%; when the CAT target is utilized, while the AUC metric remains unchanged, the GAUC metric further increases by 0.13\%. 
The comparison corroborates the effectiveness of AKA and CAT. 

\input{tables/label_ablation}

\subsubsection{Analysis of the Source of Performance Gains} \label{subsec:performance_gain}

Considering that AKA assigns a dedicated tower to each attribution CVR prediction task, this design inevitably increases the overall model parameter scale compared to the Base model. There are two possible explanations for the performance gains brought by \modelname:
\begin{itemize}[leftmargin=*] 
\item \textbf{Conjecture \Rmnum{1}}: The increase of model parameters contributes to the performance gains.
\item \textbf{Conjecture \Rmnum{2}}: The information gain from different attribution mechanisms brings performance lifts.
\end{itemize} 
To distinguish between these two conjectures, we conduct an ablation study in which all prediction towers in AKA are trained using primary targets, rather than multi-attribution signals.
As shown in Table~\ref{tab:label_ablation}, the resulting variant of \modelname becomes comparable to  Base in terms of AUC and GAUC metrics.
This indicates that simply increasing the number of parameters contributes little to the observed performance gains. 
Therefore, the effectiveness of \modelname can be largely attributed to the information gain derived from multiple attribution perspectives—supporting Conjecture \Rmnum{2}. 

\begin{figure}[!t]
\includegraphics[width=0.75\columnwidth]{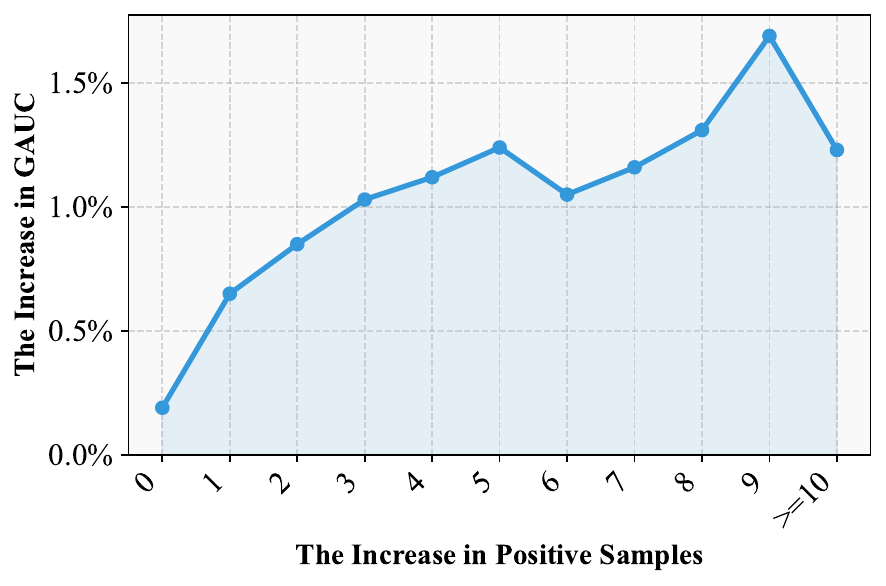}
\caption{The increase in GAUC of different user groups split by the increase in positive samples brought by multi-atttribution learning compared to the baseline. The last-click label is the primary target.}
\label{fig:gauc_by_group}
\end{figure}

To further verify Conjecture \Rmnum{2}, we group users by the information gain magnitude brought by multi-attribution learning, which is proxied by the increase in the positive sample count under linear attribution relative to last-click attribution in the last two weeks of training data, and compute GAUC within each user group.
As shown in Figure~\ref{fig:gauc_by_group}, the increase in GAUC is strongly positively correlated with the increase in positive samples.
In particular, the GAUC of the user group with zero positive sample growth only increases by \textbf{0.19\%};
in contrast, the number of the user group with a positive sample increase of no less than 10 rockets to \textbf{1.23\%} strikingly.
The finding further supports Conjecture \Rmnum{2}.

\subsubsection{Influence over Different Industries} \label{subsec:industry_analysis}
\begin{figure}[t]
\centering 
\includegraphics[width=0.75\columnwidth]{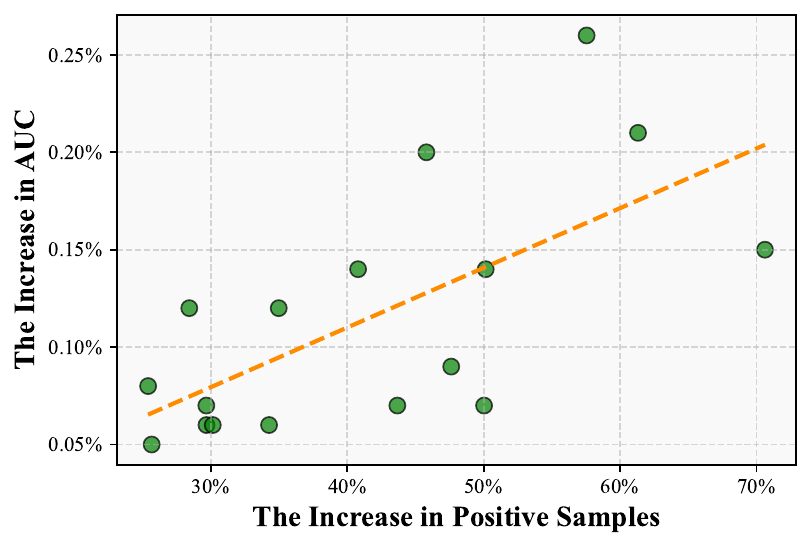}
\caption{The increase in AUC and positive samples brought by multi-attribution learning across industries.} 
\label{fig:industry_scatter}
\end{figure}

To better understand the impact of \modelname on the whole system, we group the test samples according to the advertiser's industry and compute the AUC metrics within each industry.
As plotted in Figure~\ref{fig:industry_scatter}, the AUC value increases across all industries, and the industries with higher increases in the number of positive samples achieve greater AUC growth.
In particular, as listed in Table~\ref{tab:auc_industry}, industries with long conversion paths, e.g., large home appliances and jewelry, have the highest AUC growth, while industries with short conversion paths, e.g., pet products and toys, have a significantly lower growth of AUC.
The results are consistent with the previous user-group analysis in \S~\ref{subsec:performance_gain}, indicating that MAL effectively utilizes the incremental information provided by multiple attribution mechanisms.

\input{tables/auc_by_industry}

\subsection{Online A/B Testing} 
We conduct a strict online A/B experiment in the display advertising system of Taobao, Alibaba, lasting from May 21st to May 25th, 2025.
Results show that compared to the production baseline, the \modelname model delivers an overall 2.7\% lift in GMV (Gross Merchandise Volume), a 1.2\% rise in BuyCnt (the number of orders), and a 2.6\% increase in ROI (Return on Investment).
Note that GMV, BuyCnt, and ROI are core metrics for evaluating the online performance of CVR models in our online system, and a 1\% rise in them implies a substantial improvement of advertiser benefits and platform revenues.
Therefore, the results prove that \modelname achieves better ad recommendation performance and brings significant business benefits.~\looseness=-1

Besides, we find performance differentials across industries, with advertising campaigns in high unit-price industries showing the most substantial BuyCnt lifts.
For instance, we find that large home appliances (+11.6\%) and jewelry (+9.7\%) have more pronounced BuyCnt lifts.
The reason is that these high unit-price industries exhibit long conversion paths and benefit from alternative attribution mechanisms (more holistic modeling of conversion intent). 
This phenomenon is in line with the offline analysis results in \S~\ref{subsec:industry_analysis}.~\looseness=-1

%% file: tables/main_results.tex
\begin{table}[t]
\centering
\caption{CVR Ranking performance of \modelname and baselines. All values are percentages. The best results are highlighted in bold and the second-best results are \underline{underlined}.}
\resizebox{\columnwidth}{!}{
\begin{tabular}{@{}l|cc|cc@{}}
\toprule
\textbf{Primary Target}    & \multicolumn{2}{c}{\textbf{Last-Click}} & \multicolumn{2}{|c}{\textbf{MTA}} \\ 
\cmidrule(r){1-1} \cmidrule(lr){2-3}  \cmidrule(l){4-5} 
\textbf{Method/Metrics} & \textbf{GAUC} & \multicolumn{1}{c}{\textbf{AUC}} & \multicolumn{1}{|c}{\textbf{GAUC}} & \textbf{AUC} \\ 
\midrule 
Base & 0.7819 \basex{0.00} &  0.9110 \basex{0.00} & 0.7824 \basex{0.00} &  0.9042 \basex{0.00} \\ 
Shared-Bottom~\cite{caruana1997multitask} & \underline{0.7843} \basexx{0.31} & \underline{0.9119} \basexx{0.10} & 0.7858 \basex{0.43} &  0.9053 \basex{0.12} \\ 
MMoE~\cite{mmoe}           & 0.7825 \basex{0.08} & 0.9115 \basex{0.05} & 0.7844 \basex{0.26} & 0.9049 \basex{0.08} \\ 
PLE~\cite{PLE}             & 0.7836 \basex{0.22} & 0.9117 \basex{0.08} & \underline{0.7861} \basexx{0.47} & \underline{0.9055} \basexx{0.14} \\ 
HoME~\cite{wang2024home}   & 0.7825 \basex{0.08} & 0.9116 \basex{0.07} & 0.7846 \basex{0.28} & 0.9052 \basex{0.11} \\ 
\midrule 
\textbf{\modelname (Ours)} &  \textbf{0.7859} \up{0.51} & \textbf{0.9123} \up{0.14} & \textbf{0.7873} \up{0.75} & \textbf{0.9061} \up{0.21}  \\ \bottomrule
\end{tabular}}
\label{tab:main_results}
\end{table}


%% file: tables/label_ablation.tex
\begin{table}[t]
\centering
\caption{Ablation experiment results. The last-click label is used as the primary target.} 
\resizebox{0.35\textwidth}{!}{ 
\begin{tabular}{@{}l|cc@{}} 
\toprule
\textbf{Model} & \textbf{GAUC} & \textbf{AUC} \\ \midrule
Base                                     & 0.7819 \basex{0.00}  & 0.9110  \basex{0.00} \\
\modelname w/o CAT   & 0.7849  \up{0.38}  & 0.9123 \up{0.14}   \\
\modelname w/o multiple attributions     & 0.7820 \up{0.01}     & 0.9110  \basex{0.00} \\
\modelname                               & 0.7859 \up{0.51}     & 0.9123  \up{0.13}    \\ \bottomrule 
\end{tabular}
}
\label{tab:label_ablation}
\end{table}

%% file: tables/auc_by_industry.tex
\begin{table}[t]
\centering
\caption{The AUC increase and the growth of positive samples of representative industries.}
\resizebox{0.45\textwidth}{!}{
\begin{tabular}{@{}c|lcc@{}}
\toprule
\textbf{Conversion Path} & \textbf{Industry}   & \textbf{\#Pos.} & \textbf{AUC} \\ \midrule
\multirow{3.5}{*}{Long}   & Large Home Appliances  & +70.63\%               & +0.15\% 

\\  & Sports \& Outdoors  & +61.32\%               & +0.21\%      \\
& Jewelry & +50.15\%               & +0.14\%    \\ 
                         \midrule          
\multirow{3.5}{*}{Short}   & Toys                & +34.26\%               & +0.06\%      \\
                         & Pet Products             & +30.13\%               & +0.06\%      \\
                         & Flowers & +25.39\%               & +0.08\%      \\                          
                         
                         \bottomrule
\end{tabular}} 
\label{tab:auc_industry}
\end{table}

%% file: chapters/conclusion.tex
\section{Conclusion and Discussion}
Traditional industrial CVR prediction models have predominantly relied on single-attribution learning frameworks, utilizing only the production-deployed attribution mechanism for model training. This study establishes a paradigm shift through our novel multi-attribution learning framework that strategically synthesizes signals from multiple attribution mechanisms to enhance performance under the primary attribution mechanism for production. 

Through systematic investigation of architectural design and learning objectives, we propose the \modelname framework for multiple-attribution knowledge utilization, which is further enhanced with Cartesian-based auxiliary training (CAT) for enriching the supervision signals. 
Experiments show that \modelname yields substantial improvements in CVR prediction accuracy, addressing fundamental limitations of single-attribution approaches that inadequately capture the conversion intention in user conversion journeys. Online A/B tests in the Taobao advertising system confirmed the superiority of \textbf{MAL}. The findings provide both a methodological framework and empirical insights that can support the effective integration of multi-attribution learning into CVR prediction.

For future work, we plan to extend the MAL framework by incorporating a richer spectrum of user actions beyond purchases. This will enable our modeling objectives to capture user conversion intent with greater fidelity, thereby boosting model performance.
We also believe that augmenting MAL with the powerful generative recommendation architecture~\cite{zhai2024actions} and the reasoning capability of large language models~\cite{wei2022chain} is a promising direction for renovating recommendation paradigms.